\documentclass{llncs}
\usepackage{makeidx}  %
\usepackage{etoolbox} %

\usepackage{graphicx} %
\usepackage{caption}
\usepackage[labelformat=simple]{subcaption}

\usepackage[colorinlistoftodos,prependcaption,textsize=tiny]{todonotes}
\usepackage[fixamsmath,disallowspaces]{mathtools}
\usepackage[hidelinks]{hyperref}
\usepackage{booktabs}
\makeatletter
\patchcmd{\ps@headings}{\rlap{\thepage}}{}{}{}
\patchcmd{\ps@headings}{\llap{\thepage}}{}{}{}
\makeatother
\usepackage{acro}[=v2]

\DeclareAcronym{ros}{short=ROS, long=Robot Operating System}
\DeclareAcronym{fts}{short=FTS, long=force/torque sensor}
\DeclareAcronym{mbf}{short=MBF, long=Move Base Flex}
\DeclareAcronym{navpi}{short=NavPi, long=Navigation Pipeline}
\DeclareAcronym{fzi}{short=FZI, long=FZI Forschungszentrum Informatik}
\DeclareAcronym{mca2}{short=MCA2, long=Modular Control Architecture 2}
\DeclareAcronym{hollie}{short=HoLLiE, long=House of Living Labs intelligent Escort}
\DeclareAcronym{estop}{short=e-stop, long=emergency stop}
\DeclareAcronym{dof}{short=DOF, long=degree of freedom}

\newcommand{\myac}[1]{\acifused{#1}{\ac{#1}}{\acl{#1} (\acs{#1})}}
\newcommand{\myacl}[1]{\acifused{#1}{\acl{#1}}{\acl{#1}}}
\begin{document}
\mainmatter              %
\title{HoLLiE C -- A Multifunctional Bimanual Mobile Robot Supporting Versatile Care Applications}
\titlerunning{HoLLiE C}  %

\author{Lea Steffen\inst{1} \and Martin Schulze\inst{1} \and Christian Eichmann\inst{1} \and Robin Koch\inst{1} \and Andreas Hermann\inst{2} \and Rosa Frietsch Mussulin\inst{1} \and Friedrich Graaf\inst{1} \and Robert Wilbrandt\inst{1} \and Marvin Große Besselmann\inst{1} \and Arne Roennau\inst{1} \and R\"udiger Dillmann\inst{1}}
\authorrunning{Steffen et al.} %
\institute{FZI Research Center for Information Technology, Karlsruhe, Germany,\\
\email{steffen@fzi.de} %
\and
ArtiMinds Robotics, Karlsruhe, Germany}

\maketitle              %

\begin{abstract}
Care robotics as a research field has developed a lot in recent years, driven by the rapidly increasing need for it.
However, these technologies are mostly limited to a very concrete and usually relatively simple use case.
The bimanual robot \myac{hollie} includes an omnidirectional mobile platform. This paper presents how \myac{hollie} is adapted, by flexible software and hardware modules, for different care applications. 
The design goal of HoLLiE was to be human-like but abstract enough to ensure a high level of acceptance, which is very advantageous for its use in hospitals.
After a short retrospect of previous generations of \myac{hollie}, it is highlighted how the current version is equipped with a variety of additional sensors and actuators to allow a wide range of possible applications. Then, the software stack of \myac{hollie} is depicted, with the focus on navigation and force sensitive intention recognition.

\end{abstract}
\section{INTRODUCTION}  \label{sec:introduction}
A challenge for many healthcare systems is the overburdening of caregivers and consistent understaffing in the care industry.
The healthcare system needs additional approaches, using robots could be one way to address these challenges in the healthcare system. 
Even though the human touch is deeply valued, there are still tasks that would benefit from robotic automation without a deterioration of care quality.

\subsection{Related work} \label{sec:related work}
There are already several approaches regarding robotic systems intended for care facilities and their use in hospitals~\cite{Khaksar2023}~\cite{Kyrarini2021}.
Due to an easy adoption to a healthcare setting, social robots are a prevalent robot application in care, with various systems based on commercially available robots in use~\cite{tanioka2019nursing}~\cite{Kyrarini2021}~\cite{Worlikar2022handhygiene}.
Pure logistics robots that help deliver and pick up medications, lab samples, tests, and meals, among other things, are the \emph{TUG} mobile robots from Aethon and \emph{Relay} from Swisslog Healthcare. 
They can navigate autonomously and without collisions in hospital corridors and bring objects to different locations. Unlike HoLLiE, however, they do not have a humanoid appearance and are accordingly limited in their function~\cite{Kyrarini2021}.
The Vighnaharta robot from Rucha Yantra, also not an anthropomorphic robot, can perform similar tasks. 
It consists of a mobile platform and a superstructure for specific tasks~\cite{Khaksar2023}. 
Robots that focus on non-care tasks and logistics in hospitals also include~\cite{calderon2015logistics}.
In contrast to these logistic robots, semi-humanoids, like Moxi and Pepper, with more responsibilities have also been introduced to care facilities. 
Moxi, developed by Diligent Robotics~\cite{Ackermann2020} consists of a mobile platform, a humanoid upper body with a head and a 7-\myac{dof} arm with a two-finger gripper. 
Moxi's sensors include a laser scanner and a camera. Its main tasks include picking up and distributing supplies, collecting specimens and soiled laundry bags. 
Pepper, from SoftBank Robotics, moved with the help of a mobile platform. A touch screen was located on its chest and through its arms, hands, head, hips, knees and base it has 20 \myac{dof}. In addition, the robot had several cameras, a microphone, speakers and various sensors, through which many possibilities arise. 
In addition to various applications such as an assistant in a school, the robot could also be used in hospitals. There, its tasks included working at the reception desk of a hospital and assisting hospital staff in recording health data~\cite{Kyrarini2021}.
Lastly, several robots with direct patient contact have been developed and tested in hospitals.
In Japan, a care robot is in use whose task is patient assistance. The robot with a bear-like head called ROBEAR from RIKEN and Sumitomo Riko Company Limited can carry a patient from a bed to a wheelchair. ROBEAR has a mobile platform, an upper body and two arms, and can also help the patient stand up~\cite{Kyrarini2021}. 
A robot for care facilities is the GARMI humanoid robotic assistant from Geriatronics. The robot consists of two arms, a head and an upper body, which sits on a mobile platform. 
Its tasks include helping with housework, assisting with rehabilitation exercises and telemedicine doctor visits. For this purpose, the robot is equipped with ECG, a blood pressure monitor and ultrasound~\cite{Trobinger2021}. \\
For research, KUKA, Karlsruhe University of Applied Sciences and RWTH Aachen University have launched the PeTRA hospital robot~\cite{Schule2022}. The mobile assistance robot is intended to relieve nursing staff by assisting in patient transport. 
The robot can couple to a wheelchair, a rollator or a bed and transport it throughout the facility.
In addition, the robot assists and monitors patients while walking. The robot consists of a mobile base and a collaborative robotic arm. Another research project to develop an assistance robot is HIRO (Human Interactive Robotics for Healthcare)~\cite{Khaksar2023}. 
The project develops a humanoid robot platform called EVE. 
Target tasks are to assist healthcare professionals in day-to-day tasks such as meal delivery and disinfection. In addition, the robot will accompany visitors and new patients, providing them with information as necessary. The robot has a head with a face and movable arms.
These systems for fairly general use in hospitals are supplemented by research on robotic assistance for specific diseases~\cite{rivera2019flexible,Scherzinger2202arne}.

\subsection{HoLLiEcares} \label{sec:holliecares}
The consortium of HoLLiECares\footnote{\url{https://holliecares.de/}} is composed of five technical partners\footnote{robotic startup ArtiMinds, KIT Institute for Control Systems (IRS), Fraunhofer IOSB, August-Wilhelm Scheer Institut, FZI Forschungszentrum Inforkatik}, two practice partners\footnote{Karlsruhe Municipal Hospital, Knappschaftsklinikum Saar} --- the hospitals --- with their respective nursing stations and the consortium leadership, an institute for nursing research\footnote{German Institute of Applied Nursing Research (DIP)}.
The aim is to research and further develop multi-functional service robots to support professional care in hospitals in transport, interactive assistance, and documentation.
Thereby, several use cases are developed and tested in the hospitals, \textit{a)} wound documentation, \textit{b)} exercise/physiotherapy instructions and \textit{c)} accompanying patients to examinations
\textit{d)} sorting and stocking medicines, \textit{e)} handling deformable objects~\cite{Dittus2021} and \textit{f)} pushing a wheelchair~\cite{Schulze2023}.
The intention of the project is not to replace nurses but to replace non-nursing tasks with robotic competence and provide additional functionality, to relieve the nursing staff.
To make a difference to already established specialized solutions with mostly non-complex applications, \myac{hollie}Cares focuses on multi-functionality. 
To address the multi-faceted targets of the project, the design of the robot needs to take the complexity of future nursing tasks into account and allow adaptation along the project path. 
However, this paper does not deal with the developments and results of \myac{hollie}Cares, but specifically with the adaptation of the multifunctional service robot \myac{hollie} for the hospital.

\subsection{History of the Servicerobot \myac{hollie} at FZI} \label{sec:history_hollie}
The development of the assistance robot \myac{hollie}, a bimanual mobile service and assistance robot with an actuated body, started in 2011. 
\begin{figure*}[h!]
	\centering %
	\begin{subfigure}{0.25\textwidth}
		\includegraphics[height=5cm]{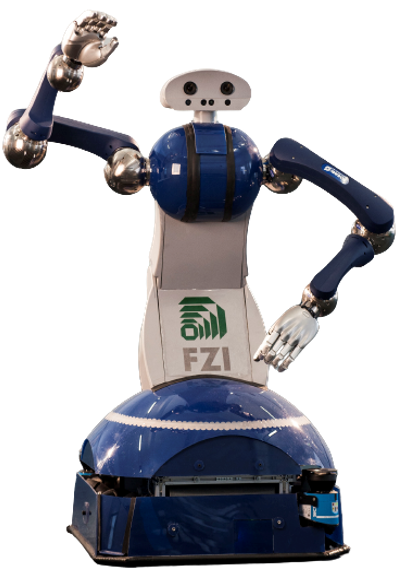}
		\caption{2012}\label{fig:hollie_1}
	\end{subfigure}\hfil %
	\begin{subfigure}{0.25\textwidth}
		\includegraphics[height=5cm]{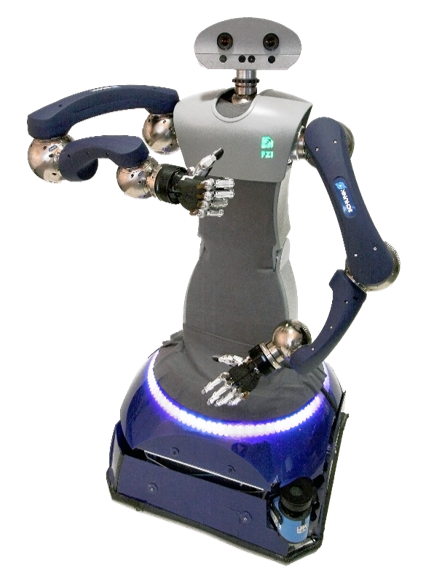}
		\caption{2013}\label{fig:hollie_2}
	\end{subfigure}\hfil %
	\begin{subfigure}{0.25\textwidth}
		\includegraphics[height=5cm]{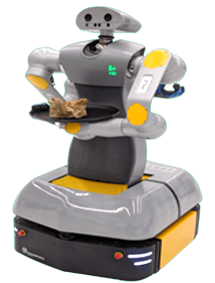}
		\caption{2019}\label{fig:hollie_3}
	\end{subfigure}\hfil %
	\begin{subfigure}{0.25\textwidth}
		\includegraphics[height=5cm]{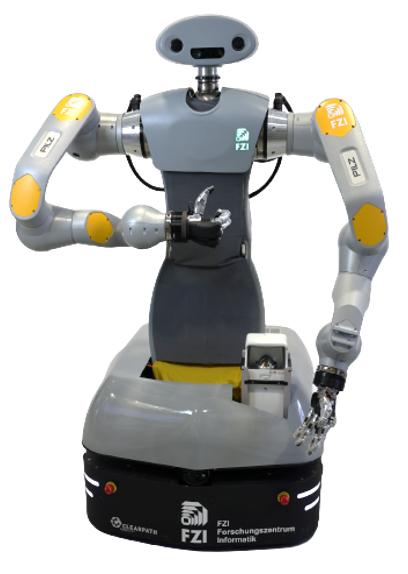}
		\caption{2023}\label{fig:hollie_4}
	\end{subfigure}\hfil %
	\caption{Development of the robot \myac{hollie} by years.  \subref{fig:hollie_1}~\cite{Hermann2013} and \subref{fig:hollie_2}~\cite{Heppner2017}.}
	\label{fig:hollie_historisch}
\end{figure*}
The FZI Forschungszentrum Informatik recognized a lack of mobile service robots that feature an articulated upper body to realize an extensive workspace for everyday manipulation tasks in human environments. 
From experience regarding past projects the following guidelines for the design of a new robot were defined~\cite{Hermann2013}:
\begin{itemize}
	\item Realize a body structure with the ability to pick up small objects from the floor. Actuators must be powerful enough to handle everyday objects.
	\item The overall design must be small and graceful with an abstract, partly anthropomorphic appearance as a non-intimidating robot body encourages people to interact.
	\item Employ commercially available hardware components wherever possible to reduce development time and guarantee reliability. Self-developed mechanisms must be robust and long-lasting while supporting extensibility and easy service.
	\item Create a closed outer hull with no parts sticking out, allowing a clean design that hides the technical complexity and creates a presentable appearance.
\end{itemize}
This hardware concept was implemented in the following year while we partnered with a product designer to shape the outer appearance of \myac{hollie} to create a friendly-looking cobot. 
The usage of fabric makes it easy to hide all mechanical body hinges while 3D printed covers and a rounded head transport flowing exterior forms.
The key concept of the body structure is a spring-loaded double parallelogram structure. 
The upper body can bow forward without affecting the orientation of the arms and neck during motions. This allows the usage of relatively small actuators while featuring a large workspace and easing hand-eye calibration.
At the same time, a small mobile base with mecanum wheels and two laser scanners enabled the robot to navigate through domestic environments with standard-sized doors.
Speakers in the torso, microphones in the head, a multi-color LED badge in the chest and an orbital multi-color LED band around the mobile base could be used to communicate with users by words but also non-verbal.
\myac{hollie} was presented to the public for the first time at the Motek fair 2012, with the outer appearance from Fig. \ref{fig:hollie_1}. 
Countless public events followed thereafter where the ability to pick up objects from the floor or to detect and mimic human motions with the built-in RGBD head camera were demonstrated. 
The software stack of \myac{hollie}'s first incarnation was based on FZI proprietary \myac{mca2}, including custom platform navigation and whole-body motion planning. 
The software was ported or replaced over time by \myac{ros} implementations~\cite{Quigley2009}. Current software is listed in Sec. \ref{sec:software}.

\section{APPROACH} \label{sec:approach}

\subsection{Hardware components} \label{sec:hardware}
\myac{hollie} was designed to allow easy maintenance of additional hardware components, as most of the installed electronics can be accessed easily when the magnetic cover on the mobile base is removed.
The removable gray hull, mounted on the ridgeback platform, can be seen in \autoref{fig:hollie_3} \& \ref{fig:hollie_4}.

\subsubsection{Computer and Power supply}
The robotic system \myac{hollie} contains three PCs, for relevant specifications see \autoref{tab:pc_specs}. 
The PCs are connected via Ethernet and use \myac{ros}~\cite{Quigley2009} to communicate with each other. 
Spreading the system onto three different PCs allows the compartmentalization of tasks in levels of abstraction. 
Low-level tasks only allow the movement of the ridgeback platform and provide the roscore, mid-level tasks enable sensor and actor communication, whereas high-level tasks consist of motion planning, navigation and mapping. The PC connected to the ridgeback handles low-level tasks. 
\begin{table}[h!]
	\centering
	\begin{tabular}{lp{4cm}p{2cm}p{4cm}}
		\toprule
						& CPU										& RAM 		& GPU \\
		\midrule
		ridgeback		& Intel(R) Core(TM) i5-4570TE CPU @ 2.70GHz	& 4 GB 		& Intel Corp. Xeon E3-1200 v3/4th \\
		hollie 			& Intel(R) Core(TM) i7-6700TE CPU @ 2.40GHz	& 16 GB 	& Intel Corp. HD Graphics 530 \\
		hollie-x-01		& Intel(R) Core(TM) i7-9700TE CPU @ 1.80GHz	& 33 GB 	& NVIDIA RTX A2000 \\
			
		\bottomrule
	\end{tabular}
	\caption{Specifications of the computers integrated into the robotic platform \myac{hollie}. Thereby, low-level tasks are executed on ridgeback, mid-level on hollie and high-level on hollie-x-01. }
	\label{tab:pc_specs}
\end{table}
The PC \textit{hollie} as well as \textit{hollie-x-01} are industrial PCs, i.e. they have robust housings, flexible DC power delivery and feature CAN and RS485 interfaces.
The \emph{hollie} PC, responsible for mid-level tasks, handles all communication with external devices such as the robot arms, the \myac{fts}, the laser scanners etc. via CAN bus, Ethernet and RS485. 
On the highest level, there is a Nuvo-7160GC by Neousys Technology Inc., which is responsible for tasks such as mapping, navigation and text-to-speech applications. 
As it is the only PC with a discrete GPU it also processes camera input.
The division into several layers allows one to only use the hardware required for a specific task while leaving the rest turned off, which helps extend the battery life span.
\begin{figure*}[h!]
	\centering
	\includegraphics[width=\textwidth]{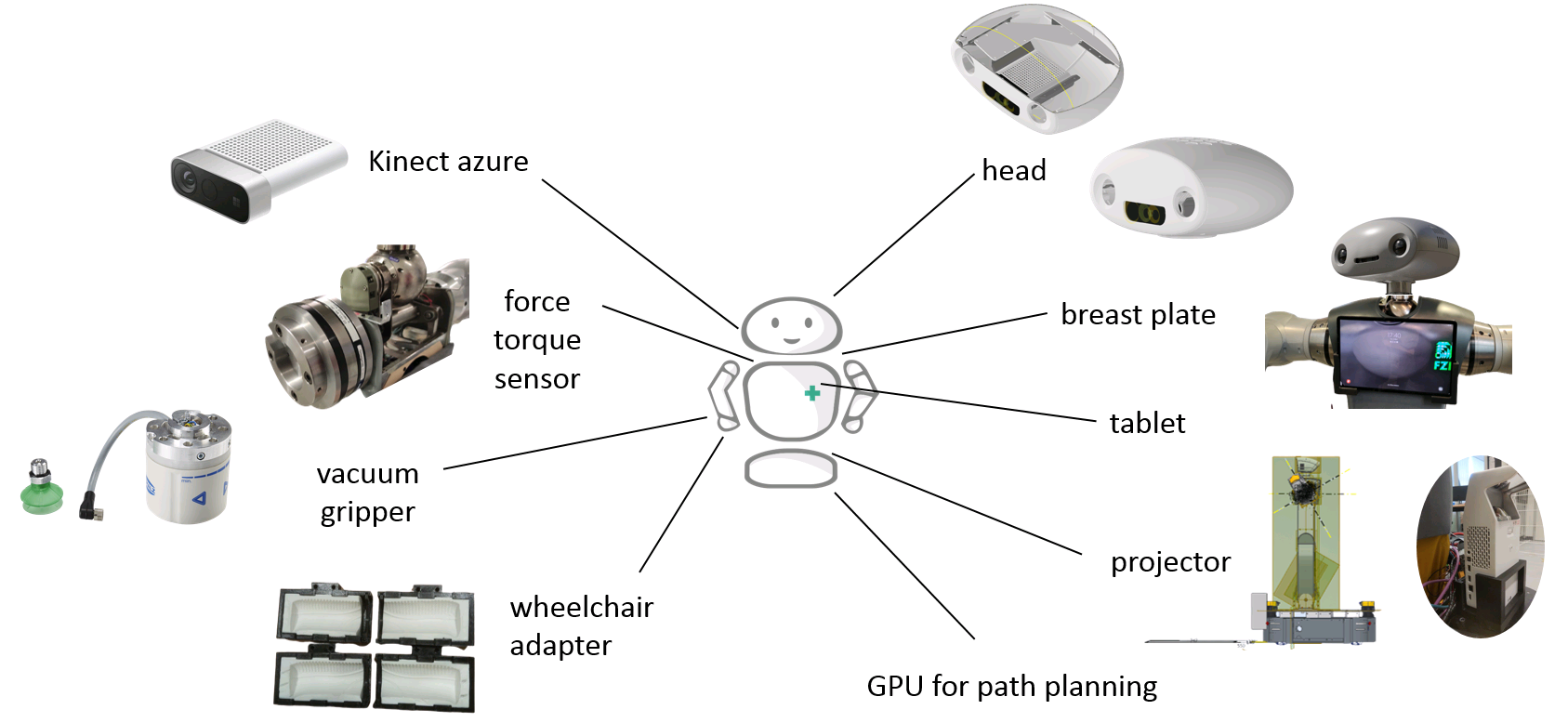}
	\caption{An overview of \myac{hollie}'s hardware components. 
	Two 6 \myac{dof} arms with easily exchangeable grippers and force moment sensors in the shoulders allow a variety of manipulation tasks. 
	A projector, a tablet, 3D cameras and speakers enable multi-modal communication.}
	\label{fig:hardware_overview}
\end{figure*}
All components are powered by two 12V 105Ah AGM batteries of the ridgeback platform, which provide around 24V that is further regulated to 5V and 12V.
Power delivery is realized through a central terminal block which is connected to the battery and placed on the ridgeback platform, to allow for easy access. 
The power delivery to the terminal can be manually switched off, allowing to run only the ridgeback platform disconnected from all other components. 
The terminal is split into two parts: a non-secured part that is always powered and a part that is connected to the \myac{estop} of the ridgeback platform. 
The non-secured part of the terminal powers all non-moving electrical components providing 5V, 12V and 24V to satisfy the various voltage needs, while the secured part of the terminal powers the 
moving components i.e. arms, end-effectors, the neck and the torso joints and provides only 24V. 
This separation of secured and non-secured power terminals ensures the safety of operation, as all movement of \myac{hollie} can be stopped by pressing one of the four \myac{estop} 
that are built into the ridgeback platform.
Wall power can be plugged and unplugged during operation and the battery run time strongly depends on the tasks performed and hardware components turned on.

\subsubsection{Actuators, Sensors and User Interfaces}
\myac{hollie} is assembled from industrial components. It has a movable torso and neck, each with two~\myac{dof}. The torso has two actuated joints in the lower and upper body and a passively driven joint in the shoulder. 
The head, which sits on a pan-tilt mount, is equipped with a 3D camera and a stereo sound system. Speakers are attached as eyes and the robot's mouth consists of a Kinect azure RGB-D camera.
\myac{hollie}'s upper body is mounted on a ridgeback, an omnidirectional mobile platform from Clearpath robotics, which is equipped with two laser scanners to detect the environment. 
\begin{figure*}[h!]
	\centering %
	\begin{subfigure}{0.24\textwidth}
		\includegraphics[width=\linewidth]{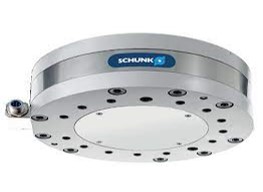}
		\caption{}\label{fig:kms}
	\end{subfigure}\hfil %
	\begin{subfigure}{0.24\textwidth}
		\includegraphics[width=\linewidth]{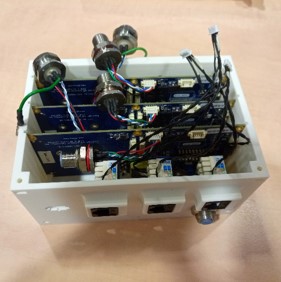}
		\caption{}\label{fig:platinen}
	\end{subfigure}\hfil %
	\begin{subfigure}{0.24\textwidth}
		\includegraphics[width=\linewidth]{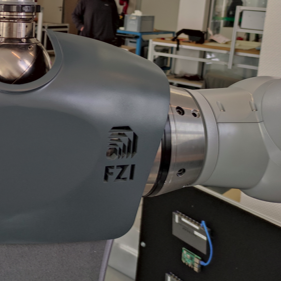}
		\caption{}\label{fig:kms_verbaut}
	\end{subfigure}
	\begin{subfigure}{0.24\textwidth}
		\includegraphics[width=\linewidth]{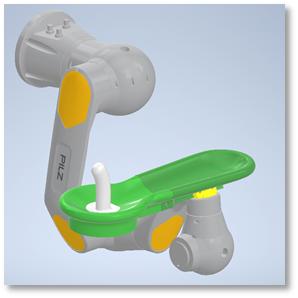}
		\caption{}\label{fig:unterarmablage}
	\end{subfigure}
	\caption{\subref{fig:kms} A \myac{fts} from Schunk. \subref{fig:platinen} A custom-made space-saving box for the circuit boards of the sensor. \subref{fig:kms_verbaut} The installed \myac{fts}. \subref{fig:unterarmablage} To provide a haptic interface between the user and the robot, a forearm rest was manufactured.}
	\label{fig:force_torque}
\end{figure*}
It also has two lightweight Pilz PRBT robot arms, each with six~\myac{dof}, with interchangeable grippers to match the specific application requirements.
Examples of grippers on \myac{hollie} are industrial jaw grippers, vacuum grippers and the  SVH 5-finger hand from Schunk, which allows grasping various common household objects~\cite{Ruehl2014}.
The robot arms had their original base removed and are now attached to the torso via an adapter plate specially designed to match the mounting of the \myac{fts} in the shoulders.
The controller PCB that was originally housed in the arm base is moved onto the ridgeback platform in a custom-made 3D-printed housing, making it easily accessible for maintenance and troubleshooting.
A \myacl{fts} from Schunk~\cite{schunk} is installed on both sides between the arms and the torso. Located in the hull of the ridgeback, there is also an embedded short throw projector that can project an image onto the floor in front of \myac{hollie}.
For multimodal communication, \myac{hollie} is additionally equipped with a tablet. The tablet holder, mounted on \myac{hollie}'s chest, allows easy removal such that the tablet can be used as either a remote control or a communication channel on \myac{hollie}.

\subsection{Software stack} \label{sec:software}
The aforementioned hardware architecture is reflected in the software stack which is also divided into 3 levels that are each run on the corresponding PC.
On every layer catmux, a tmux adaption developed at FZI\footnote{https://github.com/fmauch/catmux} which allows users to run tmux sessions which are predefined in yaml files, is used to run the different \myac{ros} nodes.
This division in combination with the usage of catmux facilitates troubleshooting since all tasks can be (re)started individually.

\subsubsection{Navigation} \label{sec:navigation}
The current navigation stack of \myac{hollie} comprises Google's Cartographer, a sensor-filter pipeline and the \myac{mbf} plugin-based framework~\cite{Ptz2018} to offer unified \myac{ros} action interfaces for planning, controlling and optional recovery behaviors. 
Moreover, \myac{mbf} allows global planners and local controllers to be switched during run time and is compatible with previous Move Base plugins. 
Currently, we are relying on a Dijkstra global planner implementation offered by the \myac{ros} \texttt{nav\_core} package. 
Depending on the current use-case for \myac{hollie}, we either load a monolithic local controller plugin (e.g. timed-elastic-bands) or the FZI's \myac{navpi} plugin. 
The \myac{navpi} offers a modularized local controller architecture and plugins for commonly shared controller responsibilities including path-sampling, obstacle avoidance, safety margins, local planning and generating steering commands. \\
For free and unconstrained point-to-point navigation, \myac{hollie} currently utilizes the \myac{navpi} using a simple path-sampling approach similar to the \myac{ros} BaseLocalPlanner, 
no recovery behavior, and simple 2D-costmap path-following with reactive obstacle avoidance solely based on 360-degree laser scans. 
In constrained navigation scenarios with few exchangeable behaviors, such as coupled wheelchair navigation~\cite {Schulze2023}, the local controllers are loaded as regular \myac{mbf} compatible controllers. 
Filtering sensor data for both, camera and laser scanners, is implemented via the \myac{ros} \texttt{sensor\_filters} package\footnote{\url{http://wiki.ros.org/sensor_filters}}. 
The filter pipeline for the laser scanners reduces computational complexity by coarsening the resulting point cloud, removing outliers and putting a lower bound on the accepted distance for each ray to prevent dust on the sensors from being detected as obstacles.

\subsubsection{Digital Twin} \label{sec:digital_twin}
\begin{figure*}[h!]
	\centering
	\includegraphics[width=0.6\textwidth]{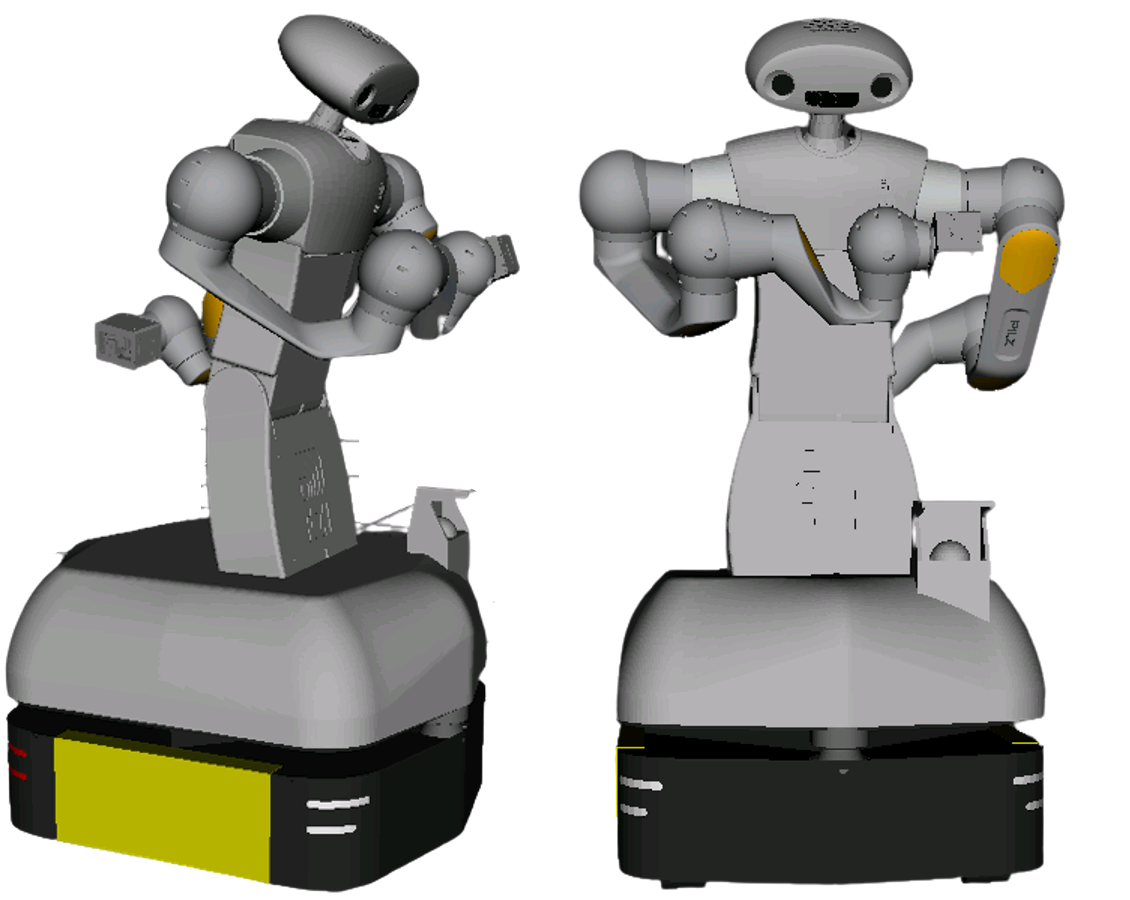}
	\caption{Visualization of \myac{hollie}'s digital twin. }
	\label{fig:hollie_sim}
\end{figure*}

The virtual twin represents a digital model based on a physical one -- in this case the robot platform \myac{hollie}. 
In the course of \myac{hollie}Cares a virtual twin was realized in the form of Docker containers, containing the original control algorithms and a Gazebo simulation via a detailed 3D model as visualized in \autoref{fig:hollie_sim}. 
Besides the simulation, the virtual twin allows visualizing and working with live data in a digital model.
In this way, all of \myac{hollie}'s joint angle positions can be mapped continuously on the simulated model, which has many advantages, especially for motion planning.
Another major advantage is protecting the hardware while testing new features. Furthermore, it enables software developers to co-develop a system without being tied to the hardware on-site~\cite {Singh2021Twin}.
Consequently, the development of a digital twin transforms \myac{hollie} into a great platform for research and technical developments, which is well-suited for collaborating with external partners.
Prototypes for grippers can be tested before they are manufactured, which is cheaper, more time efficient and reduces waste. In addition, problems during motion sequences can be predicted, increasing safety. 
Within the project \myac{hollie}Cares it allowed partners to prepare the integration offline in simulation. 
Thereby, being able to test motion sequences on \myac{hollie} without being on-site, which significantly reduced the integration time on the physical system. 
Lastly, there is no safety risk when working with the digital twin, while working with \myac{hollie} requires more caution.

\subsubsection{Force Sensitive Control} \label{sec:force_sensitive_control}
Installing the \myac{fts} allows the heavy robot \myac{hollie} to be moved by gently pulling and pressing on its arms.
To do this, the perceived force is mapped to a speed specification. 
The armrest shown in \autoref{fig:unterarmablage} provides a comfortable user interface for this purpose. However, \myac{hollie}'s arm can also be pushed or pulled directly. 
The forces and torques are measured using the \myac{fts} in \myac{hollie}'s shoulder, located between each arm and the torso, as visualized in \autoref{fig:kms_verbaut}. 
To remove the static force induced by the mass of the robot arm, the sensors are tared beforehand. 
From these measurements, only the forces are transferred to the respective shoulder's coordinate system. The torques are omitted for multiple reasons. 
For once, their correct assessment is dependent on the exact interaction point between the robot and the human. 
They also provide no added benefit for deriving the intended translation to the robot base." \\
After transfer into the robot coordinate system, the force $F$ is scaled, whereby the sign is retained and an upper and lower limit is introduced. 
This ensures that steady-state sensor noise does not lead to the platform moving without any user input. 
Hence, the lower limit enables robustness against sensor noise and the upper limit prevents extreme reactions for large forces. 
The former is necessary as otherwise the system is constantly disturbed by minimal deflections due to vibrations. 
The latter ensures that the misconduct of the users does not lead to dangerous situations.
Subsequently, the scaled forces are mapped linearly on the target speed $[-0,5; 0,5] \frac{m}{s}$.
How the 3D vector $F$ is scaled and subsequently mapped on a target speed is formalized as:
\begin{equation}
	f(x) = \begin{cases}
		\frac{F}{|F|} \times 0.5 \frac{m}{s} & \text{for } |F| \geq 120 N     \\
		0 \frac{m}{s}                        & \text{for } |F| \leq 20 N      \\
		\frac{F}{120} \times 0.5 \frac{m}{s} & \text{for } 20 N < |F| < 120 N \\
	\end{cases}
\end{equation}
Before the robot platform is controlled, this value is smoothed by
\begin{equation}
	0.98 \times speed_{old} + 0.02 \times speed_{new}
\end{equation}
This is necessary due to the severe wiggling in the upper body during motion. Hence, smoothing the original 100 Hz signal prevents a feedback loop.
However, this form of post-processing creates a certain inertia but increases the safety and reliability of the approach, which has a high priority for the usage of \myac{hollie}.
\section{RESULTS AND EXPERIMENTS} \label{sec:experiments}
\begin{figure}[h!]
	\centering
	\begin{subfigure}{.48\textwidth}
		\centering
		\includegraphics[height=4cm]{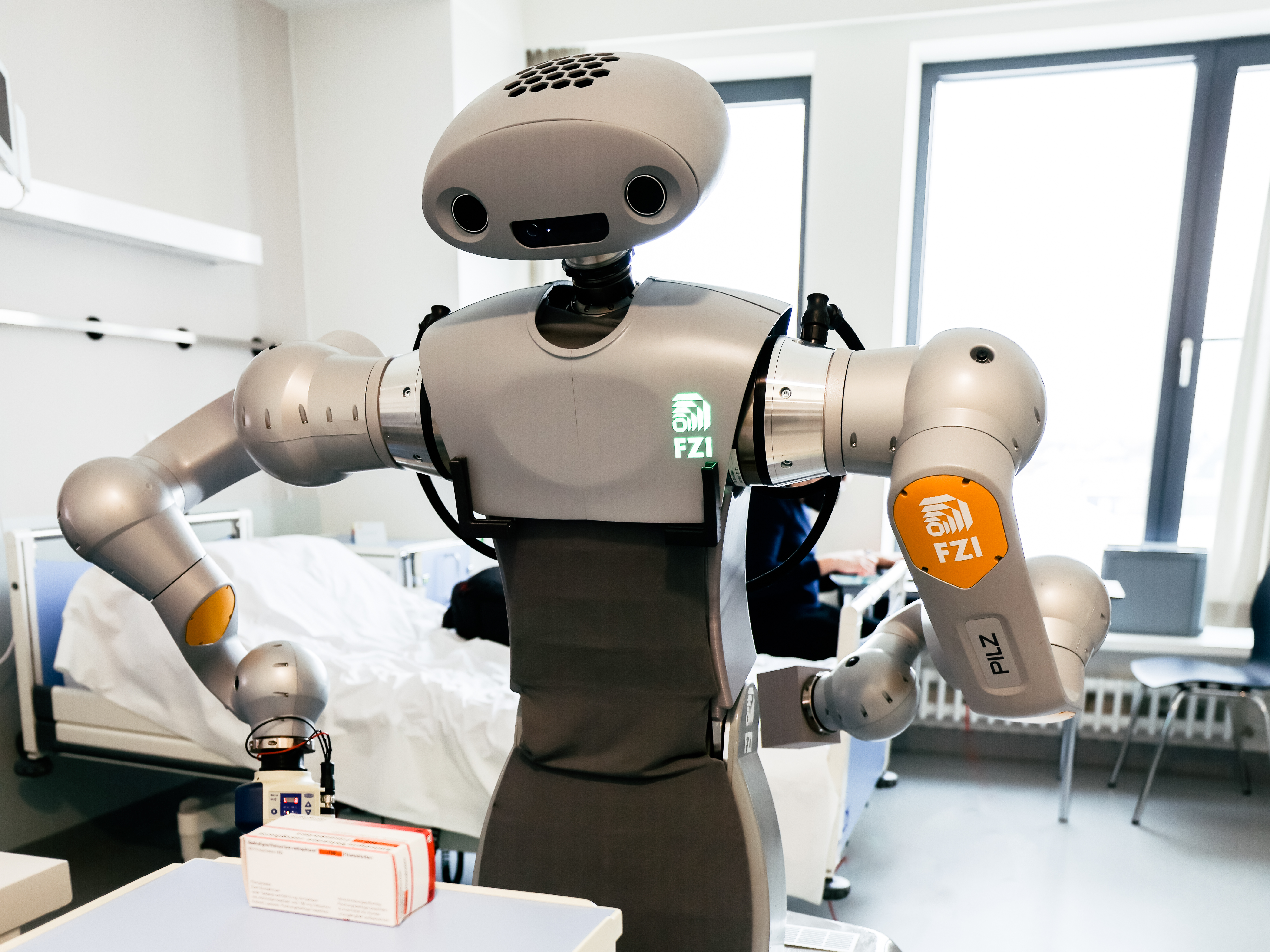}
		\caption{}
		\label{fig:hollie_medis}
	\end{subfigure}
	\hspace{0.2cm}
	\begin{subfigure}{.48\textwidth}
		\centering
		\includegraphics[height=4cm]{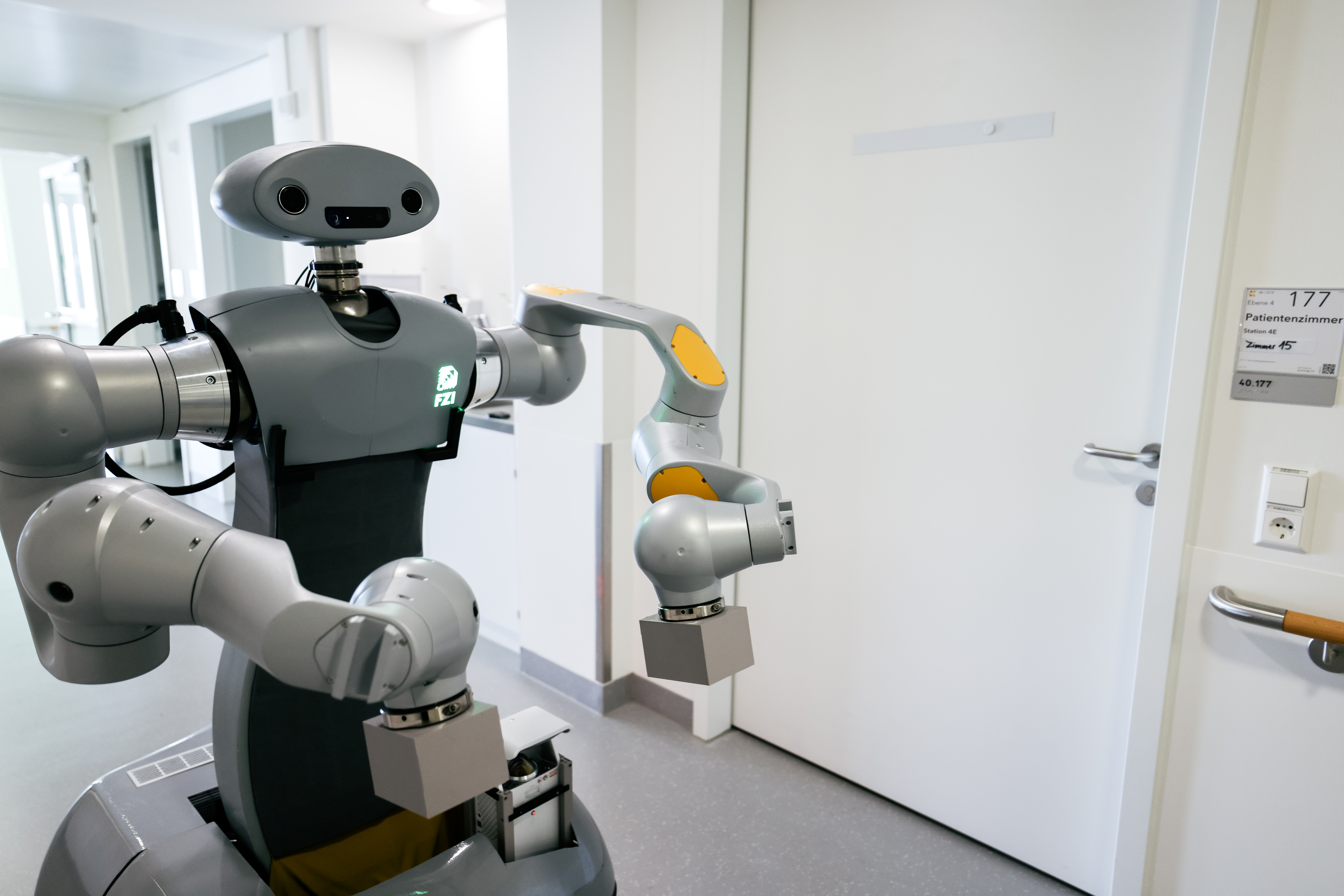}
		\caption{}
		\label{fig:hollie_close}
	\end{subfigure}%
	\caption{\myac{hollie} during a practical test phase in the Karlsruhe Municipal Hospital.}
	\label{fig:hollie_hospital}
\end{figure}

The project \myac{hollie}Cares included two practical test phases carried out for one week in each hospital, as visualized in \autoref{fig:hollie_hospital}.
Thereby, the navigation stack was evaluated in the hospital and we collected feedback from patients and nursing staff on the robotic system. 
The most significant criticism we encountered from patients and nursing staff members was the actual size of \myac{hollie}'s hardware but especially the dimensions of the ridgeback platform.
It was difficult to maneuver through doors, due to the shoulder-to-shoulder width of the upper body, and if the corridor was blocked by beds or trolleys, human help was required. 
Patients and staff had expected a smaller system that is more compatible with the hospital environment. 
The reactions of the nursing staff were predominantly positive and were sometimes accompanied by technical interest, especially among younger staff members.
However, some nurses feared that humanity and empathy would be lost, especially in the care of elderly people.
Some other reactions stated a fear of losing their job in the foreseeable future.
The reactions of patients were also mostly positive. 
Although HoLLiE was perceived as too bulky, positive feedback came for the design. 
The robot was seen as friendly and did not scare users. The robotic system was perceived as solid and creating confidence.

\subsection{Navigation} \label{sec:eval_navigation}
\begin{figure}[h!]
	\centering
	\begin{subfigure}{.48\textwidth}
		\centering
		\includegraphics[height=5.4cm]{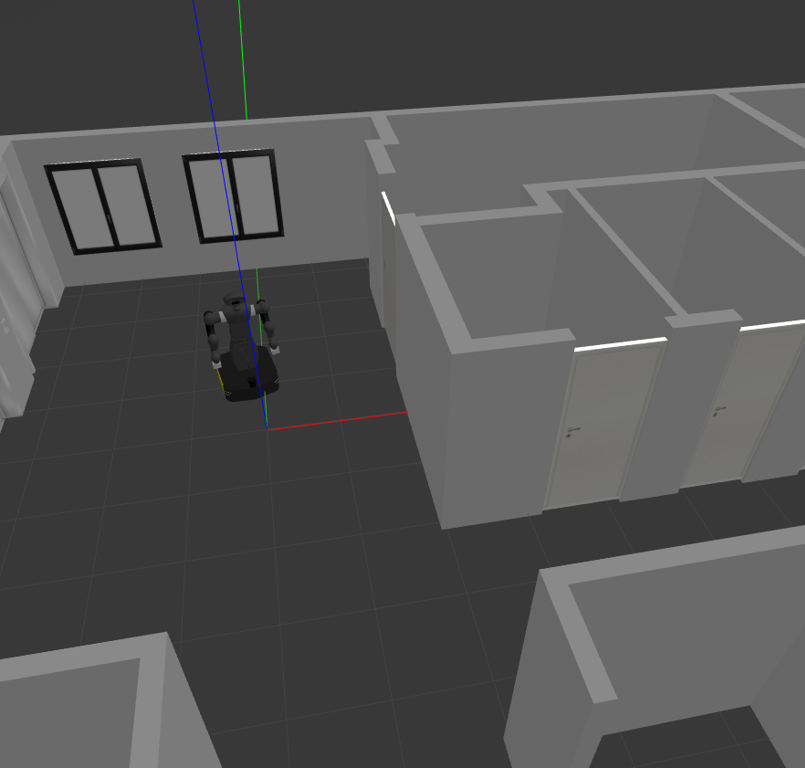}
		\caption{}
		\label{fig:hollie_gaz}
	\end{subfigure}
	\hspace{0.2cm}
	\begin{subfigure}{.48\textwidth}
		\centering
		\includegraphics[height=5.4cm]{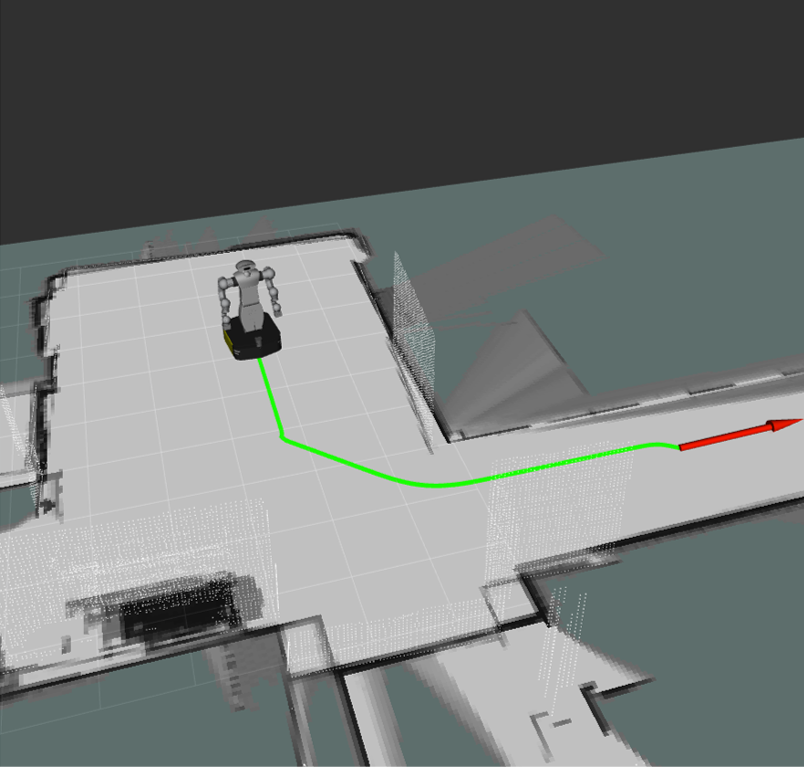}
		\caption{}
		\label{fig:hollie_map}
	\end{subfigure}%
	\caption{\subref{fig:hollie_gaz} A simulation of \myac{hollie} in a map based on a floor of the Karlsruhe Municipal Hospital. 
	\subref{fig:hollie_map} Generated map and the global path, marked in green, line towards a goal, visualized as a red arrow.}
	\label{fig:hollie_navigation}
\end{figure}
The result of the navigation implementation in simulation can be seen in \autoref{fig:hollie_navigation}.
The picture shows the global plan as created by using the Dijkstra algorithm on the 2D cost map, that \myac{hollie} is following by utilizing the 2D path following algorithm.
The small white dots show the filtered sensor data from the Kinect azure camera and the lidar sensors. Based on this a 2D cost map is built up and updated and the robot \myac{hollie} localizes itself in it.
With the 2D path following algorithm being capable of evading obstacles unseen by the global plan, the computational cost to recalculate the global plan for every new obstacle found is greatly reduced.
By shifting the executed path into a viable area, away from obstacles and walls, the robustness against disturbances like localization drift and inaccuracies is increased.
Even with \myac{hollie} having a platform capable of omnidirectional movement we actively do not use this capability for regular movements, as this would make the movements more difficult to anticipate for 
people around \myac{hollie}, and thus reduce the acceptance of the system in a normal working environment.
Recording a complete map during evaluation in the hospitals, represented as a 2D occupancy grid, did turn out to not be trivial as even in late hours of the day 
robotic systems drag very much the attention of curious patients, which often like to gather around the platform, do not move and end up being marked as permanent 
obstacles in the resulting map.
Another practical limitation arises from our path planning algorithms and their parametrization which currently do not allow \myac{hollie} to drive sideways 
through door frames which is necessary as the shoulder-to-shoulder width of \myac{hollie}'s upper body could not pass some doors to patients' rooms in the hospitals.

\subsection{Force Sensitive Control} \label{sec:eval_force_sensitive_control}
\begin{figure}[h!]
	\centering
	\begin{subfigure}{.48\textwidth}
		\centering
		\includegraphics[height=3.5cm]{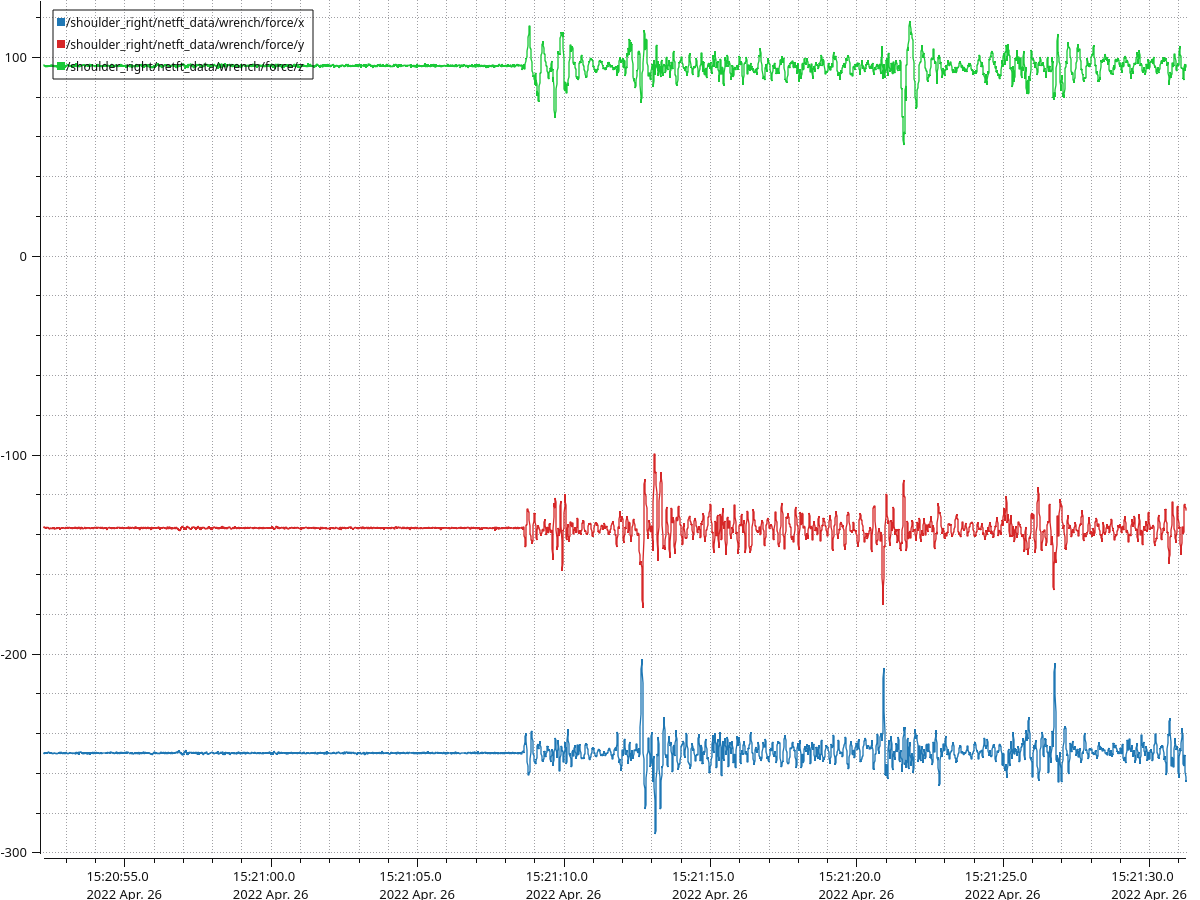}
		\caption{}
		\label{fig:C_Forces_without_TARA}
	\end{subfigure}
	\hspace{0.2cm}
	\begin{subfigure}{.48\textwidth}
		\centering
		\includegraphics[height=3.5cm]{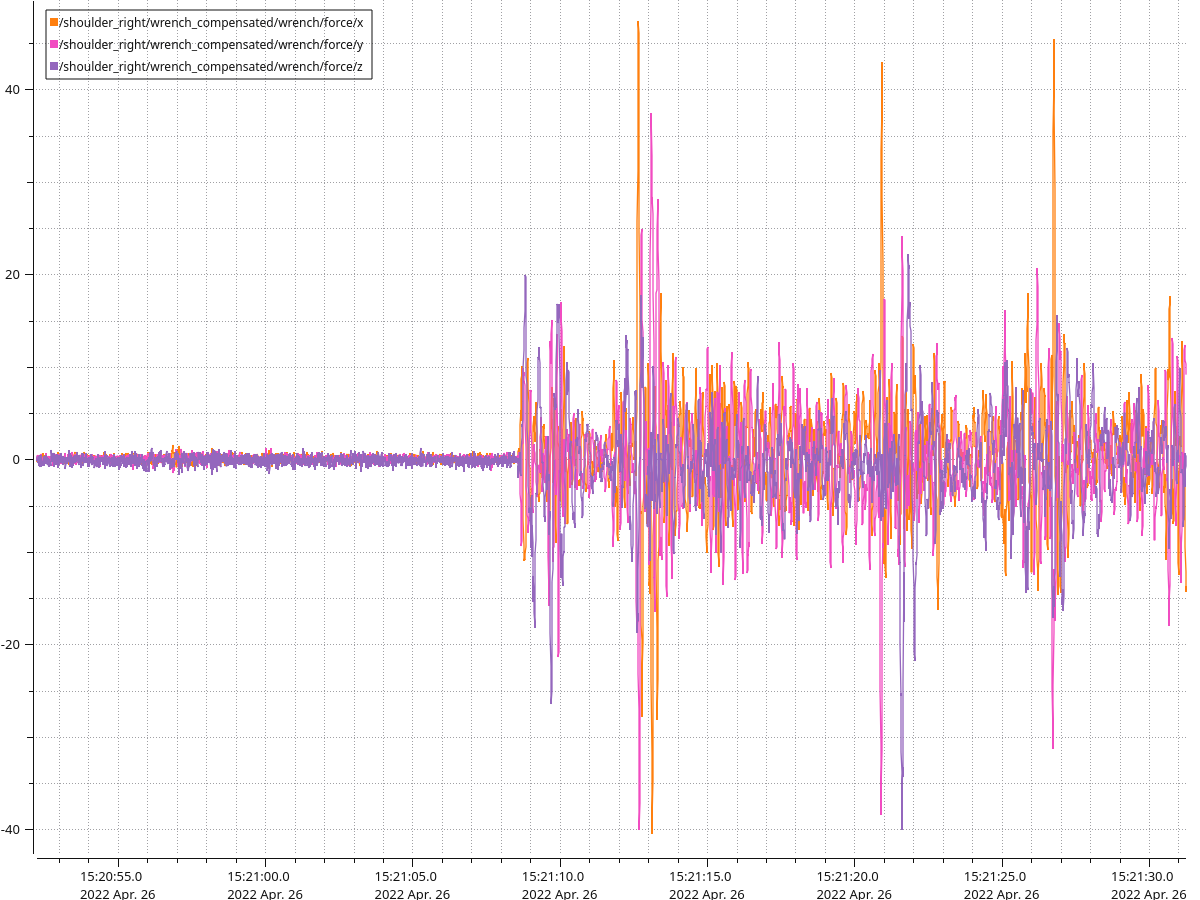}
		\caption{}
		\label{fig:C_Forces_with_TARA}
	\end{subfigure}%
	\hspace{0.2cm}
	\begin{subfigure}{.48\textwidth}
		\centering
		\includegraphics[height=3.5cm]{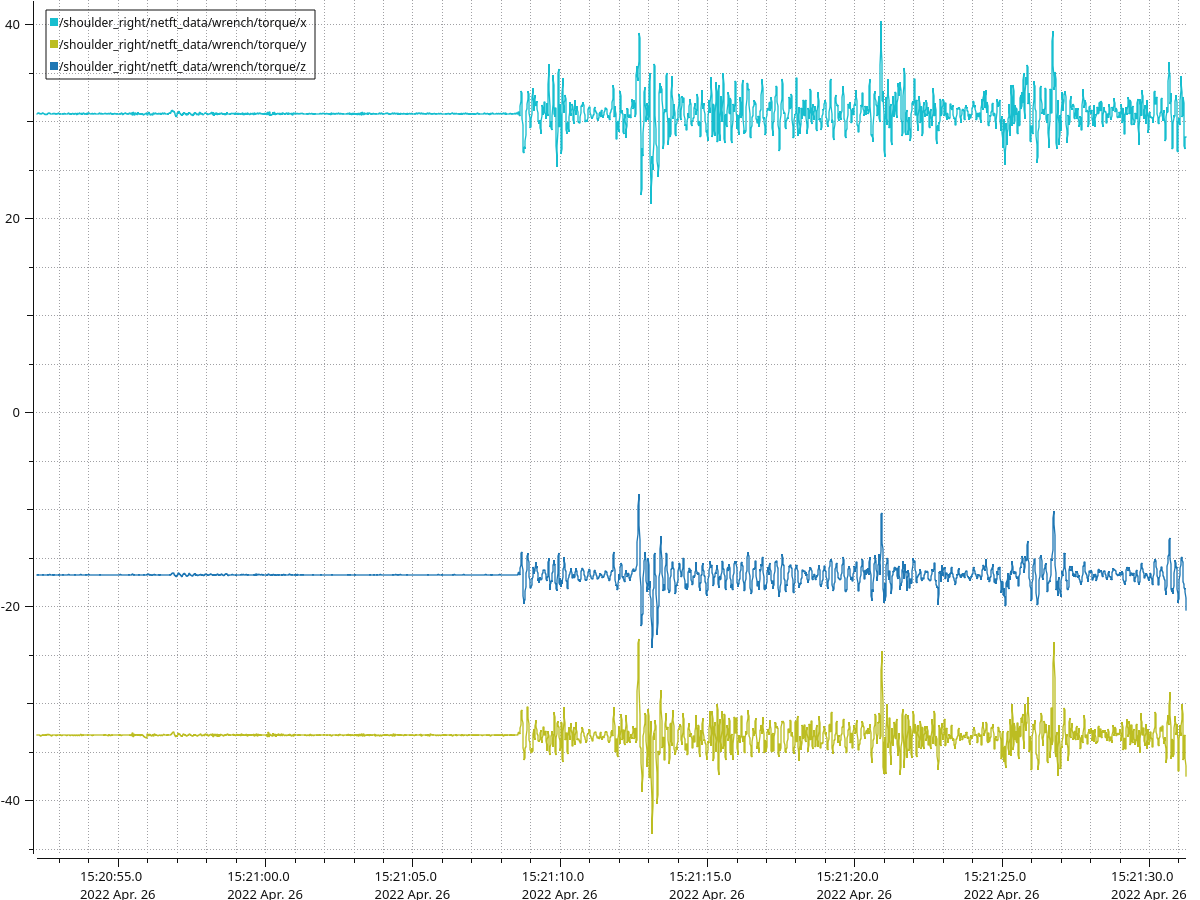}
		\caption{}
		\label{fig:C_Torques_without_TARA}
	\end{subfigure}
	\hspace{0.2cm}
	\begin{subfigure}{.48\textwidth}
		\centering
		\includegraphics[height=3.5cm]{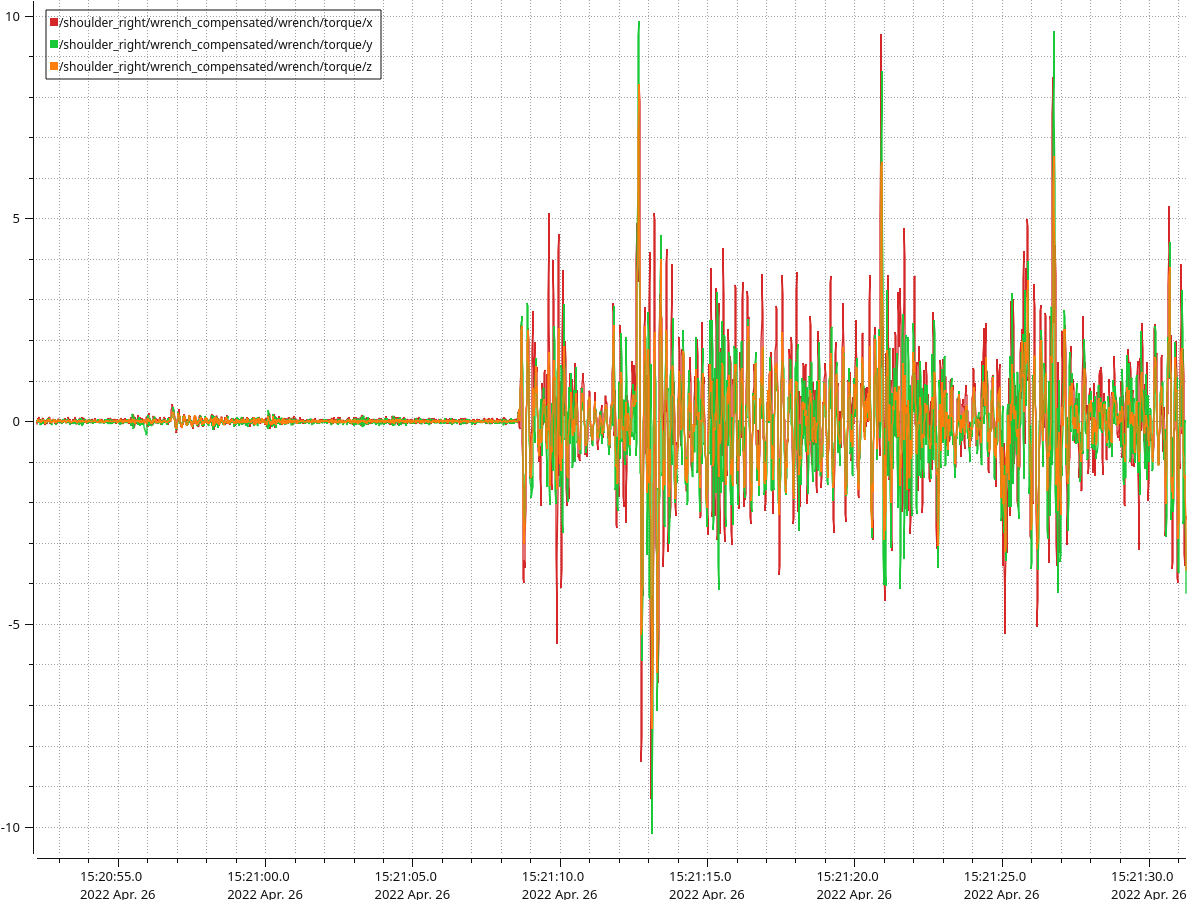}
		\caption{}
		\label{fig:C_Torques_with_TARA}
	\end{subfigure}
	\caption{Impact of the tara service on forces and torques. \subref{fig:C_Forces_without_TARA} raw data, \subref{fig:C_Forces_with_TARA} cleaned data regarding forces. \subref{fig:C_Torques_without_TARA} raw data, \subref{fig:C_Torques_with_TARA} cleaned data regarding torques.}
	\label{fig:C_tareing_service}
\end{figure}

A simple evaluation of the functionality of the force-sensitive system on \myac{hollie} is the determination of a force delta, as shown in \autoref{fig:C_tareing_service}.
Due to the relatively high net weight of the arms and their exposed position, a large disturbing force acts on the \myac{fts}. 
However, only the force delta that indicates the actual force exerted on the arm is of interest.
\begin{figure*}[h!]
	\centering
	\includegraphics[width=0.5\textwidth]{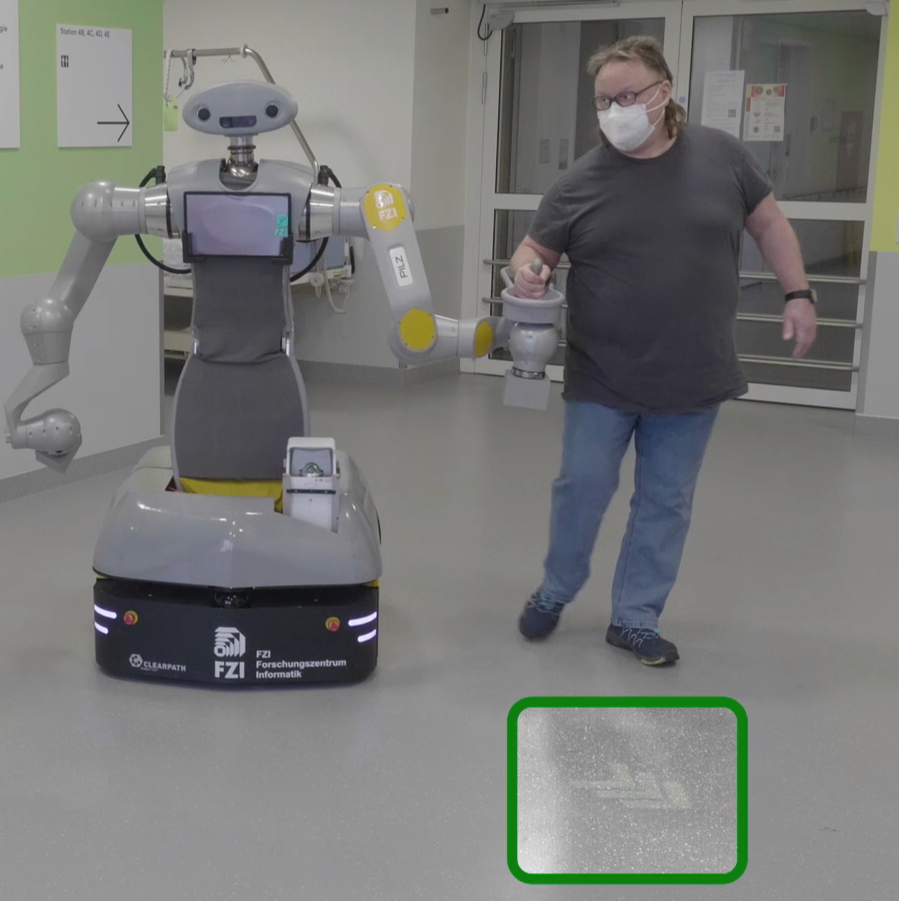}
	\caption{Visualization of the recognition of user intent implemented using \myac{fts} in \myac{hollie}'s shoulders. The direction in which the user wants to move \myac{hollie} is determined and the result is displayed via the projector and simultaneously transferred to the platform as motion commands.}
	\label{fig:fts_arrow}
\end{figure*}
To obtain this value, a \myac{ros}-service\footnote{\url{https://github.com/ipa320/g_compensator}} was used which stores the current value to offset all future measurements. 
Live calculations are not required as the arm pose is constant when the robot is pulled or pushed.
In \autoref{fig:C_Forces_with_TARA} \& \ref{fig:C_Torques_without_TARA} the adjusted values are shown and additionally in \autoref{fig:C_Forces_with_TARA} and \ref{fig:C_Forces_with_TARA} unadjusted force and torque measurements are provided as reference values. 
To evaluate the system described in \autoref{sec:force_sensitive_control}, a visualization of the assessed direction was integrated. To increase the user-friendliness of the application, this is visible to any user using an integrated projector, as shown in \autoref{fig:fts_arrow}.

\section{FUTURE WORK}  \label{sec:conclusion}
In the course of \myac{hollie}Cares, several use cases have been developed and evaluated in hospitals. Respective results are to be published as future work.
Regarding navigation, semantically annotated maps should be integrated to allow more flexibility and autonomy.
Furthermore, it has also been shown that it is necessary to develop a sophisticated communication dialog system. Which is an interesting focus for future projects.

\section*{Acknowledgement}
This research received funding as the project HoLLiEcares from the German Federal Ministry of Education and Research (BMBF) under grant no. 16SV8406.

\bibliographystyle{splncs}
\bibliography{bibliography}

\end{document}